\section{Methodology}
\label{sec:methodology}

\subsection{Model}
The central idea in BagPack (\textbf{Bag}-of-words representation of
\textbf{Pa}ired \textbf{c}oncept \textbf{k}nowledge) is to construct a
vector-based representation of a pair of words in such a way that the
vector represents both the contexts where the two words co-occur and
the contexts where the single words occur on their own. A
straightforward approach is to construct three different sub-vectors,
one for the first word, one for the second word, and one for the
co-occurring pair. The concatenation of these three sub-vectors is the
final vector that represents the pair.

This approach provides us a graceful fall back mechanism in case of
data scarcity. Even if the two words are not observed co-occurring in
the corpus -- no syntagmaic information about the pair --, the
corresponding vector will still represent the individual contexts
where the words are observed on their own. Our hypothesis (and hope)
is that this information will be representative of the semantic
relation between the pair, in the sense that, given pairs
characterized by same relation, there should be paradigmatic
similarity across the first, resp.~second elements of the pairs (e.g.,
if the relation is between professionals and the typical tool of their
trade, it is reasonable to expect that that both professionals and
tools will tend to share similar contexts).

Before going into further details, we need to describe what a
``co-occurrence'' precisely means, define
the notion of context, and determine how to structure our vector. For
a single word $W$, the following pseudo regular expression
identifies an observation of \textit{occurrence}: \begin{eqnarray} &
``C\;W \;D"\label{deneme}\end{eqnarray} where $C$ and $D$ can be empty strings or concatenations of up to 4 words separated by whitespace (i.e. $C_1,\dots,C_i$ and $D_1,\dots,D_j$ where $i,j\le4$). Each observation of this pattern constitutes a \textit{single context} of $W$. The pattern is
matched with the longest possible substring without crossing sentence
boundaries.

Let $(W_{1},W_{2})$ denote an ordered pair of words $W_{1}$ and
$W_{2}$. We say the two words \textit{occur as a pair} whenever one of the
following pseudo regular expressions is observed in the
corpus: \begin{eqnarray}
  & ``C\;W_{1}\;D\,W_{2}\;E"\label{co-occur1}\\
  &
  ``C\:W_{2}\:D\:W_{1}\:E"\label{co-occur2}\end{eqnarray}
where $C$ and $E$ can be empty strings or concatenations of up to 2 words and similarly, $D$ can be either an empty string or concatenation of up to 5 words (i.e. $C_1,\dots,C_i$, $D_1,\dots,D_j$, and $E_1,\dots,E_k$ where $i,j\le2$ and $k\le5$). Together, patterns 2 and 3 constitute the \emph{pair context} for $W_1$ and $W_2$. The pattern is
matched with the longest possible substring while making sure that $D$
does not contain neither $W_{1}$ nor $W_{2}$.

The number of context words allowed before, after, and between the
targets are actually model parameters but for the experiments reported
in this study, we used the aforementioned values with no attempt at
tuning.

The vector representing $(W_{1},W_{2})$ is a concatenation $\mathbf{v_{1}v_{2}v_{1,2}}$, %
%
%
where, the sub-vectors $\mathbf{v_{1}}$ and $\mathbf{v_{2}}$ are
constructed by using the single contexts of $W_{1}$ and $W_{2}$
correspondingly (i.e. by pattern ~\ref{deneme}) and the sub-vector
$\mathbf{v_{1,2}}$ is built by using the pair contexts identified by
the patterns \ref{co-occur1} and \ref{co-occur2}. We refer to the
components as \emph{single-occurrence vectors} and
\emph{pair-occurrence vector} respectively.

The population of BagPack starts by identifying the $b$ most frequent
unigrams and the $b$ most frequent bigrams as \textit{basis
  terms}. Let $T$ denote a basis term. For the construction of
$\mathbf{v_{1}}$, we create two features for each term $T$: $t_{pre}$
corresponds to the number of observations of $T$ in the single
contexts of $W_1$ occurring before $W_1$ and $t_{post}$ corresponds to
the number of observations of $T$ in the single occurrence of $W_1$
where $T$ occurs after $W_1$ (i.e. number of observations of the
pattern \ref{deneme} where $T\in C$ and $T\in D$ correspondingly). The
construction of $\mathbf{v_{2}}$ is identical except that this time
the features correspond to the number of times the basis term is
observed before and after the target word $W_2$ in single
contexts. The construction of the pair-occurrence sub-vector
$\mathbf{v_{1,2}}$ proceeds in a similar fashion but in addition, we
incorporate also the order of $W_{1}$ and $W_{2}$ as they co-occur in
the pair context: The number of observations of the pair contexts
where $W_{1}$ occurs before $W_{2}$ and $T$ precedes (follows) the
pair, are represented by feature $t_{+pre}$ ($t_{+post}$). The number
of cases where the basis term is in between the target words is
represented by $t_{+betw}$. The number of cases where $W_{2}$ occurs
before $W_{1}$ and $T$ precedes the pair is represented by the feature
$t_{-pre}$. Similarly the number of cases where $T$ follows (is in
between) the pair is represented by the feature $t_{-post}$
($t_{-betw}$).

Assume that the words "only" and "that" are our basis terms and
consider the following context for the word pair ("cat",~"lion"): "Lion is
the only cat that lives in large social groups." The observation of
the basis terms should contribute to the pair-occurrence sub-vector
$\mathbf{v_{1,2}}$ and since the target words occur in reverse order,
this context results in the incrementation of the features
$only_{-betw}$ and $that_{-post}$ by one.

To sum up, we have $2b$ basis terms ($b$ unigrams and $b$ bigrams).
Each of the single-occurrence sub-vectors $\mathbf{v_{1}}$ and
$\mathbf{v_{2}}$ consists of $4b$ features: Each basis term gives rise
to 2 features incorporating the relative position of basis term with
respect to the single word. The pair-occurrence sub-vector,
$\mathbf{v_{1,2}}$, consists of $12b$ features: Each basis term gives
rise to $6$ new features; $\times3$ for possible relative positions of
the basis term with respect to the pair and $\times2$ for the order of
the words. Importantly, the $2b$ basis terms are picked only once, and
the overall co-occurrence matrix is built once and for all for
\emph{all} the tasks: unlike Turney, we do not need to go back to the
corpus to pick basis terms and collect separate statistics for
different tasks.

The specifics of the adaptation to each task will be detailed in Section~\ref{sec:tasks}. For the moment, it should suffice to note that the
vectors $\mathbf{v_{1}}$ and $\mathbf{v_{2}}$ represent the contexts
in which the two words occur on their own, thus encode paradigmatic
information. However, $\mathbf{v_{1,2}}$ represents the contexts in
which the two words co-occur, thus encode sytagmatic information.

The model training and evaluation is done in a 10-fold
cross-validation setting whenever applicable. The reported performance
measures are the averages over all folds and the confidence intervals
are calculated by using the distribution of fold-specific results. The
only exception to this setting is the SAT analogy questions task
simply because we consider each question as a separate mini dataset
as described in Section \ref{sec:tasks}.

\subsection{Source Corpora}

We carried out our tests on two different corpora: ukWaC, a
Web-derived, POS-tagged and lemmatized collection of about 2 billion
tokens,\footnote{\url{http://wacky.sslmit.unibo.it}} and the Yahoo!
database queried via the BOSS
service.\footnote{\url{http://developer.yahoo.com/search/boss/}} We
will refer to these corpora as ukWaC and Yahoo from now on.

In ukWaC, we limited the number of occurrence and co-occurrence
queries to the first 5000 observations for computational
efficiency. Since we collect corpus statistics at the lemma level, we
construct Yahoo! queries using disjunctions of inflected forms that
were automatically generated with the NodeBox Linguistics
library.\footnote{\url{http://nodebox.net/code/index.php/Linguistics}}
For example, the query to look for ``lion'' and ``cat'' with 4 words
in the middle is: ``(lion OR lions) {*} {*} {*} {*} (cat OR cats OR
catting OR catted)''. Each pair requires 14 Yahoo! queries (one for
$W_1$, one for $W_2$, 6 for $(W_1,W_2)$, in that order, with 0-to-5
intervening words, 6 analogous queries for $(W_2,W_1)$). Yahoo!
returns maximally 1,000 snippets per query, and the latter are
lemmatized with the
TreeTagger\footnote{\url{http://www.ims.uni-stuttgart.de/projekte/corplex/TreeTagger/}}
before feature extraction.

\subsection{Model implementation}\label{sub:Model-realization}

We did not carry out a search for ``good'' parameter values. Instead, the model parameters are generally picked
at convenience to ease memory requirements and computational
efficiency. For instance, in all experiments, $b$ is set to 1500
unless noted otherwise in order to fit the vectors of all pairs at our
hand into the computer memory.

Once we construct the vectors for a set of word pairs, we get a
\emph{co-occurrence matrix} with pairs on the rows and the features on
the columns. In all of our experiments, the same normalization method
and classification algorithm is used with the default parameters:
First, a TF-IDF feature weighting is applied to the co-occurrence
matrix \cite{Salton:Buckley:1988}. Then following the suggestion of
Hsu and Chang (2003), each feature $t$'s
$[\hat{\mu}_{t}-2\hat{\sigma}_{t},\hat{\mu}_{t}+2\hat{\sigma}_{t}]$
interval is scaled to $[0,1]$, trimming the exceeding values from
upper and lower bounds (the symbols $\hat{\mu}_{t}$ and
$\hat{\sigma}_{t}$ denote the average and standard deviation of the
feature values respectively). For the classification algorithm, we use
the C-SVM classifier and for regression the $\epsilon$-SVM regressor,
both implemented in the Matlab toolbox of \newcite{Canu:etal:2005}. We
employed a linear kernel. The cost parameter C is set to 1 for all
experiments; for the regressor, $\epsilon=0.2$. For other pattern
recognition related coding (e.g., cross validation, scaling, etc.) we
made use of the Matlab PRTools \cite{Duin:2001}.

For each task that will be defined in the next section, we evaluated
our algorithm on the following representations: 1) Single-occurrence
vectors ($\mathbf{v_{1}v_{2}}$ condition) 2) Pair-occurrence vectors
($\mathbf{v_{1,2}}$ condition) 3) Entire co-occurrence matrix
($\mathbf{v_{1}}\mathbf{v_{2}}\mathbf{v_{1,2}}$ condition).